\documentclass{fairmeta}
\usepackage{amsmath}
\usepackage{enumerate} 
\usepackage{bm}
\usepackage{algorithm}
\usepackage{floatflt}
\usepackage{algpseudocode}
\usepackage{amsfonts}
\usepackage{amsthm}
\usepackage{newtxtt}
\usepackage{algorithm}
\usepackage{colortbl} 
\usepackage{cleveref}
\usepackage{diagbox} 
\usepackage[utf8]{inputenc}
\usepackage{textgreek}
\usepackage{colortbl}
\usepackage{nicematrix}
\usepackage{makecell}
\usepackage{float}
\usepackage{arydshln}
\usepackage[frozencache,cachedir=.]{minted}
\usepackage{caption}
\usepackage{subcaption}
\usepackage{tcolorbox}
\usepackage{amssymb}
\usepackage{xspace}
\usepackage{wrapfig}
\usepackage{adjustbox}
\usepackage{tabularx}
\usepackage{booktabs}
\usepackage{mathtools}
\usepackage{wrapfig}
\usepackage{amssymb}
\usepackage{graphicx}

\usepackage{silence}
\makeatletter
\patchcmd{\wrong@fontshape}{\@gobbletwo}{}{}{}
\makeatother
\definecolor{upColor}{RGB}{17,138,21}
\definecolor{downColor}{RGB}{174,36,67}

\newtheorem{theorem}{Theorem}[]

\newtheorem{remark1}[theorem]{Remark}

\title{Privacy-Aware Camera 2.0  Technical Report}
\author[]{Huan Song}
\author[]{Shuyu Tian}
\author[]{Ting Long}
\author[]{Jiang Liu}
\author[]{Cheng Yuan}
\author[]{Zhenyu Jia}
\author[]{Jiawei Shao}
\author[]{Xuelong Li}
\affiliation[]{Institute of Artificial Intelligence (TeleAI), China Telecom}
\metadata[Correspondence to]{Xuelong Li (\email{xuelong\_li@ieee.org})}

\begin{document}

\abstract{
With the increasing deployment of intelligent sensing technologies in highly sensitive environments such as restrooms and locker rooms, visual surveillance systems face a profound privacy–security paradox. Existing privacy-preserving approaches, including physical desensitization, encryption, and obfuscation, often compromise semantic understanding or fail to ensure mathematically provable irreversibility. Although Privacy Camera 1.0 eliminated visual data at the source to prevent leakage, it provided only textual judgments, leading to evidentiary blind spots in disputes. To address these limitations, this paper proposes a novel privacy-preserving perception framework based on the AI Flow paradigm and a collaborative edge–cloud architecture. By deploying a visual desensitizer at the edge, raw images are transformed in real time into abstract feature vectors through nonlinear mapping and stochastic noise injection under the Information Bottleneck principle, ensuring identity-sensitive information is stripped and original images are mathematically unreconstructable. The abstract representations are transmitted to the cloud for behavior recognition and semantic reconstruction via a “dynamic contour” visual language, achieving a critical balance between perception and privacy while enabling illustrative visual reference without exposing raw images.
}

\maketitle

\section{Introduction}
As intelligent sensing technology increasingly penetrates high-privacy environments such as restrooms, changing rooms, and hospital wards, visual surveillance systems are confronted with a severe privacy-security paradox \citep{song2026real,sidibe2025privacy}. On one hand, these concealed spaces are frequently hotspots for safety hazards, including falls, smoking, and school bullying. On the other hand, the public holds strong psychological resistance and ethical concerns regarding being gazed at, recorded, and stored in such settings. For a long time, these locations have remained blind spots in security regulation, trapping managers in a binary dilemma of either sacrificing privacy for safety or abandoning perception entirely to protect personal privacy \citep{an2026single,ge2023passive}. 

To circumvent the privacy controversies associated with raw RGB images, researchers initially introduced non-visual sensors, such as Thermal Sensor Arrays (TSA) and Time-of-Flight (ToF) detection \citep{newaz2024low,ofodile2019action}. While these alternatives discard identity exposure by avoiding facial texture acquisition, they suffer from a severe "Semantic Gap". The lack of texture details makes it exceedingly difficult for these sensors to accurately identify fine-grained behaviors like smoking or minor physical conflicts. Even Event Cameras, which only record pixel brightness changes, fail to guarantee absolute mathematical irreversibility against reconstruction attacks. 

Returning to visual processing, traditional image obfuscation techniques (such as blurring or pixelation) face an irreconcilable contradiction between privacy protection and data utility, often leading to a sharp decline in downstream task accuracy. More critically, recent deep learning attacks can easily penetrate these protective layers to recover original faces, rendering traditional obfuscation ineffective against strong adversarial AI. Advanced cryptographic solutions like Federated Learning and Homomorphic Encryption, while secure, impose extremely high demands on computing power and bandwidth, restricting their large-scale, real-time deployment. Furthermore, early solutions like "Privacy Camera 1.0" attempted to protect privacy by completely cutting off the video feed at the source, leaving managers with only cold, text-based alerts such as "Detected suspected fighting behavior". When a real conflict occurs, relying solely on text inference makes it difficult to determine the nature of the event, leaving the truth trapped in an information blind spot without illustrative visual reference.

To break this technical deadlock, we propose a novel privacy-preserving perception technology based on the AI Flow \citep{shao2025ai,an2026ai} theoretical framework and an edge-cloud collaborative architecture \citep{yuan2025task,liang2025integrating}. Under this architecture, the edge camera is no longer a traditional video recorder but acts as a sketch artist. In the millisecond-level instant of data acquisition, the edge device leverages a constrained feature learning algorithm based on Information Bottleneck theory to transform raw images into abstract feature vectors via non-linear mapping and random noise injection \citep{wu2025multi,shao2021learning}. This process constructs a unidirectional information flow that forcibly and irreversibly filters out identity-sensitive attributes, such as facial features and clothing, right at the physical source. Consequently, only highly lightweight feature vectors are transmitted across the network. Even if the transmission link is illegally intercepted, hackers will only obtain discrete digital trajectories from which reconstructing the original image or identifying the individual is mathematically impossible. 

When these abstract feature vectors arrive at the cloud, multimodal family models take over the joint inference. Instead of merely outputting a text-based result, the cloud utilizes homologous models to reconstruct the vectors into dynamic contour animations. This approach successfully achieves the orthogonal decoupling of semantic understanding and identity information, ensuring that what managers see is not a specific individual, but rather the truth of the behavior. Whether it is clearly displaying the frequency and force of pushing in a bullying incident, precisely capturing the regular pattern of smoking, or instantly determining whether an elderly person is slowly sitting down or suddenly falling in a care facility, these contour images allow the actions to speak for themselves. This paradigm upgrade is not just a visual transformation, but the perfection of the evidence-gathering loop. It transforms the privacy camera from a simple sensor into a more trustworthy "digital witness," helping it monitor safely while minimizing intrusion.

\section{Privacy-Preserving Edge-Cloud Collaborative Architecture}
\begin{figure}[H]
    \centering
    \includegraphics[width=0.75\linewidth]{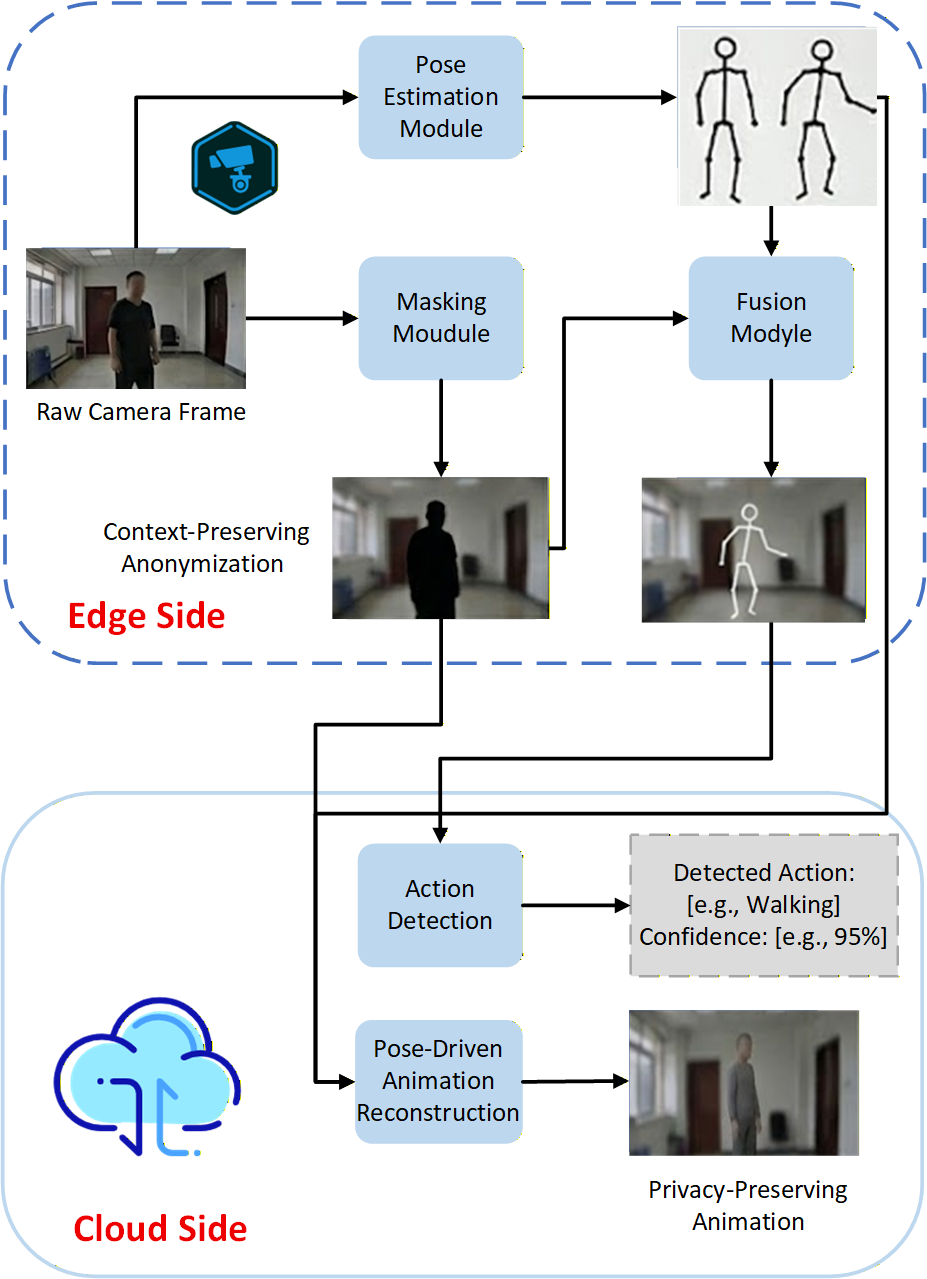}
    \caption{The overall framework of the proposed edge-cloud collaborative privacy-preserving architecture.}
    \label{fig:placeholder}
\end{figure}

This section proposes an edge-cloud collaborative technical scheme for privacy-sensitive scenarios, strictly adhering to the principle of ``Data utility without visibility'': human-related raw visual pixels are utilized exclusively at the edge for one-time feature extraction, and are subsequently physically eliminated to prevent their transmission over wide-area networks and storage in the cloud. As illustrated in Fig. \ref{fig:placeholder}, the system adopts a three-stage pipeline architecture: the edge perception module performs object detection and temporal tracking at the edge to determine the Region of Interest (ROI) and subject identifiers, and extracts pose and behavioral representations within the ROI. Furthermore, instance segmentation is utilized to obtain precise masks, erasing the pixels corresponding to the ROI from the frame to generate desensitized frames. Concurrently, the desensitized subject representations and environmental context are visually encoded to obtain compact vector representations. The secure transmission link only transmits environmental information, behavioral vectors, and visual encoding vectors, excluding reversible identity appearance pixels. Finally, the cloud reasoning and reconstruction module receives the aforementioned representations, leverages large vision models for joint reasoning and scene reconstruction, and synchronously outputs behavior recognition results and anonymized reconstructed scenes, thereby maintaining scene comprehensibility and operational utility without exposing identity details. 
\subsection{Edge Perception Module}
This module accomplishes object locking, desensitization, and vectorized representation extraction on real-time video streams at the edge. Taking the frame-by-frame image $I_t \in \mathbb{R}^{H \times W \times 3}$ as input, the system first localizes targets and generates Regions of Interest (ROI) through object detection and temporal tracking, while assigning a SubjectID to each independent individual to maintain cross-frame consistency and trajectory continuity. For the $i$-th target, denoting its geometric region as $b_t^i$, pose estimation is performed within the ROI to extract $K$ body keypoints and their confidences:
\begin{equation}
P_t^i=\{(u_{t,k}^i, v_{t,k}^i, \rho_{t,k}^i)\}_{k=1}^{K}.
\end{equation}
This set of keypoint coordinates is independently recorded as the pose estimation output.

To obtain representations that possess robust behavioral semantics and are easily processed by visual encoders, this study does not directly compress the keypoint sequence into abstract vectors, but rather maps it into the skeletal topology of an ``anthropomorphic proxy''. Specifically, combining the pose keypoint set $P_t^i$ of the $i$-th target with the optional head orientation geometric parameter $h_t^i$, the appearance representation is constructed through proxy rendering mapping:
\begin{equation}
A_t^i = \Psi(P_t^i, h_t^i, b_t^i).
\end{equation}
This generates a skeletal proxy based on the ROI coordinate system.

Simultaneously, within the joint mask $m_t=\bigvee_i m_t^i$ region generated by instance segmentation, irreversible pixel-level erasure is performed on the original image to extract a pristine environmental background (Environment). This background is saved in PNG image format as the second independent output:
\begin{equation}
\bar{I}_t = \mathcal{E}(I_t, m_t).
\end{equation}

On this basis, all target proxies are superimposed back onto the desensitized background $\bar{I}_t$ according to their geometric spatial relationships, generating an anonymized synthesized image:
\begin{equation}
\tilde{I}_t = \mathrm{Overlay}\!\left(\bar{I}_t,\ \{A_t^i\}_{i=1}^{N_t};\ \pi_t\right),
\end{equation}
where $\pi_t$ represents the occlusion relationships and rendering order among multiple targets. To avoid generating blurred pixel blocks when target spaces overlap, the parameter $\pi_t$ is dynamically inferred using the temporal trajectory prior extracted by the DeepSORT algorithm, thereby ensuring that the behavioral model generated in complex interactive scenarios possesses strict trajectory continuity.

Finally, the visual encoder encodes this anonymous synthesized frame to obtain the third core output of this module---the Vision Embedding:
\begin{equation}
z_t^{\mathrm{vis}} = f_\theta(\tilde{I}_t).
\end{equation}
This embedding effectively fuses the environmental context and anonymized behavioral cues, and together with the environmental images and pose coordinates, it supports the subsequent high-dimensional semantic reasoning tasks of the cloud-based large models.

\subsection{Secure Transmission Link}
To adapt to the input requirements of Vision-Language Models (VLMs) within the edge-cloud architecture while strictly safeguarding privacy, this stage abstracts the transmission link into a privacy-secure multimodal information bottleneck. The de-identified representation stream output by the edge perception module consists of the purified environment image $\bar{I}_t$, the set of target pose parameters $\mathcal{P}_t = \{P_t^i\}_{i=1}^{N_t}$, and the anonymous visual semantic embedding $z_t^{\mathrm{vis}}$. To ensure that the cloud-based VLM can perform accurate joint  reasoning, the system introduces a unified synchronization key $\kappa_t$ to enforce strict frame-level consistency constraints:
\begin{equation}
\kappa_t = \langle \mathrm{camera\_id},\ \mathrm{frame\_id},\ t \rangle
\end{equation}

At any time $t$, the environmental background, pose parameters, and high-dimensional visual semantic embeddings are bound via $\kappa_t$ into a strictly aligned representation tuple:
\begin{equation}
\Omega_t = \{\kappa_t,\ \bar{I}_t,\ \mathcal{P}_t,\ z_t^{\mathrm{vis}}\}
\end{equation}

This mechanism achieves two core objectives at the data flow level. First, it maintains the strict correspondence of multi-source representations through $\kappa_t$, providing reliable input for the cross-modal fusion of the VLM. Second, the representation tuple $\Omega_t$ thoroughly filters out pixels containing the original appearance and reversible textures of individuals, retaining only high-level semantic and structural information. This physically blocks the risk of exposing biometric features, achieving privacy-secure delivery to the cloud.

\subsection{Cloud Reasoning and Reconstruction Module}
The cloud reasoning module receives the de-identified representation tuple $\Omega_t$ and feeds it into the Cloud Large Foundation Model $\mathcal{M}$ for fusion reasoning. This reasoning process independently generates structured semantic results:
\begin{equation}
(R_t, \mathcal{A}_t, \mathcal{C}_t) = \mathcal{M}(\bar{I}_t, \mathcal{P}_t, z_t^{\mathrm{vis}}),
\end{equation}
where $R_t$ represents the object detection and recognition results, $\mathcal{A}_t$ denotes the fine-grained behavioral semantic labels, and $\mathcal{C}_t$ is the confidence evaluation of the corresponding output.

In the dimension of scene reconstruction, the system adopts a two-stage strategy of ``representation restoration to visual generation'' to achieve anonymized visual presentation. First, based on the pose parameters $\mathcal{P}_t$, a proxy rendering mapping is executed to restore the discrete keypoint coordinates into a Skeletal Proxy Image with dynamic characteristics:
\begin{equation}
A'_t = \Psi(\mathcal{P}_t, \pi_t).
\end{equation}
This image represents the human posture using skeletal topology and simplified geometric primitives, effectively masking sensitive appearance and identity features. Based on this, the environmental image $\bar{I}_t$ and the generated skeletal proxy image $A'_t$ are jointly input into the Visual Generative Model $\mathcal{G}$, utilizing generative priors to reconstruct the instantaneous scene:
\begin{equation}
\hat{I}_t = \mathcal{G}(\bar{I}_t, A'_t).
\end{equation}
The generated $\hat{I}_t$ ensures full-link privacy desensitization while maintaining the authenticity of the environmental context and the accuracy of behavioral postures. Ultimately, the cloud achieves the synchronous delivery of standardized behavioral semantics $(\mathcal{A}_t, \mathcal{C}_t)$ and the restored scene $\hat{I}_t$, providing a semantic reference that is both privacy-compliant and intuitively understandable for decision support.

\section{Conclusion}
To address privacy-sensitive scenarios, this paper proposes an edge-cloud collaborative architecture governed by the "data utility without visibility" principle. The framework operates through a three-stage pipeline: edge perception, secure transmission, and cloud inference. At the edge, the system detects targets, extracts pose keypoints to form a skeletal proxy, irreversibly scrubs human pixels to extract a clean background, and generates a compact visual embedding. A secure transmission link then streams only these desensitized representations synchronously, ensuring no raw appearance pixels or identifiable biometrics enter the network. Finally, cloud-based foundation models perform joint semantic inference for behavior recognition and utilize generative models to reconstruct anonymized scenes. Ultimately, this approach delivers standardized behavioral semantics and privacy-preserving visualizations, achieving high scene intelligibility and operational utility without compromising identity details. 

\bibliographystyle{plainnat}
\bibliography{paper}

\end{document}